\documentclass[conference]{IEEEtran}
\IEEEoverridecommandlockouts

\usepackage{cite}
\usepackage{amsmath,amssymb,amsfonts}
\usepackage{algorithm}
\usepackage{algpseudocode}
\usepackage{graphicx}
\usepackage{textcomp}
\usepackage{xcolor}
\usepackage{float}
\usepackage{booktabs}
\usepackage{multirow}
\usepackage{array}
\usepackage{subfigure}
\usepackage{url}
\usepackage{hyperref}
\usepackage{listings}
\usepackage{tikz}
\usepackage{pgfplots}
\usepackage{afterpage}

\pgfplotsset{compat=1.16}

\newcolumntype{C}[1]{>{\centering\arraybackslash}p{#1}}

\usepackage{tcolorbox}
\tcbuselibrary{skins,breakable}

\definecolor{myblue}{RGB}{0,124,204}
\definecolor{mygreen}{RGB}{40,167,69}
\definecolor{myred}{RGB}{220,53,69}
\definecolor{mygray}{RGB}{108,117,125}

\begin{document}

\title{Breast Cell Segmentation Under Extreme Data Constraints: Quantum Enhancement Meets Adaptive Loss Stabilization}

\author{
    \IEEEauthorblockN{Varun Kumar Dasoju}
    \IEEEauthorblockA{
        Department of Computer Science \\
        University of Wisconsin Milwaukee \\
        Milwaukee, WI 53201, USA \\
        Email: vdasoju@uwm.edu
    }
        \and
    \IEEEauthorblockN{Qingsu Cheng}
    \IEEEauthorblockA{
        Department of Biomedical Engineering \\
        University of Wisconsin Milwaukee \\
        Milwaukee, WI 53201, USA \\
        Email: chengq@uwm.edu
    }
    \and
    \IEEEauthorblockN{Zeyun Yu}
    \IEEEauthorblockA{
        Department of Computer Science \\
        University of Wisconsin Milwaukee \\
        Milwaukee, WI 53201, USA \\
        Email: yuz@uwm.edu
    }
}

\maketitle

\begin{abstract}

\textit{Annotating medical images demands significant time and expertise, often requiring pathologists to invest hundreds of hours in labeling mammary epithelial nuclei datasets. We address this critical challenge by achieving 95.5\% Dice score using just 599 training images for breast cell segmentation, where just 4\% of pixels represent breast tissue and 60\% of images contain no breast regions. Our framework uses quantum-inspired edge enhancement via multi-scale Gabor filters creating a fourth input channel, enhancing boundary detection where inter-annotator variations reach ±3 pixels. We present a stabilized multi-component loss function that integrates adaptive Dice loss with boundary-aware terms and automatic positive weighting to effectively address severe class imbalance, where mammary epithelial cell regions comprise only 0.1\%–20\% of the total image area. Additionally, a complexity-based weighted sampling strategy is introduced to prioritize the challenging mammary epithelial cell regions. The model employs an EfficientNet-B7/UNet++ architecture with a 4-to-3 channel projection, enabling the use of pretrained weights despite limited medical imaging data. Finally, robust validation is achieved through exponential moving averaging and statistical outlier detection, ensuring reliable performance estimates on a small validation set (129 images). Our framework achieves a Dice score of 95.5\% ± 0.3\% and an IoU of 91.2\% ± 0.4\%. Notably, quantum-based enhancement contributes to a 2.1\% improvement in boundary accuracy, while weighted sampling increases small lesion detection by 3.8\%. By achieving groundbreaking performance with limited annotations, our approach significantly reduces the medical expert time required for dataset creation, addressing a fundamental bottleneck
in clinical perception AI development.}
\end{abstract}

Index terms - Breast cancer segmentation, quantum-inspired computing, weighted sampling, medical imaging, deep learning, UNet++, limited annotations, class imbalance

\section{Introduction}

Breast cancer affects 2.3 million women globally each year, making it the most diagnosed cancer worldwide and emphasizing the critical need for accurate automated segmentation in medical imaging \cite{who2023cancer}. While deep learning has revolutionized medical image analysis, achieving clinical-grade accuracy ($>$95\% Dice score) remains challenging due to subtle tissue boundaries, extreme class imbalance, and limited annotated data. Current state-of-the-art methods achieve approximately 94\% Dice score \cite{isensee2021nnu}, but the remaining 6\% often represents small lesions and boundary regions crucial for early detection. However, segmentation remains challenging due to tumor heterogeneity in size, morphology, and intensity, along with multifocal presentations in up to 40\% of cases. Annotation costs further constrain dataset size, expert delineation takes 15–30 minutes per scan as documented in benchmark datasets\cite{mcclymont2014duke}, requires double review, and totals over \$26,000 for 87 volumes at specialist rates. These clinical and economic constraints motivate our focus on maximizing segmentation performance with limited data; achieving a 95.5\% Dice score with only 599 training images demonstrates that careful model design can substantially mitigate data scarcity in medical AI.

Our investigation into pushing segmentation accuracy beyond 94\% revealed unexpected challenges that fundamentally question current training practices for high-performance medical imaging models. During initial experiments, we observed catastrophic validation failures where performance would suddenly drop from 92\% to as low as 13.66\% after stable training for dozens of epochs. These failures, previously unreported in breast cancer segmentation literature, posed a significant barrier to achieving our target of 95\% Dice score.

The extreme class imbalance in our dataset, where cancer pixels represent less than 4\% of all pixels, creates a challenging optimization landscape where small gradient perturbations can cause dramatic performance degradation. Traditional approaches using complex multi-component loss functions \cite{lin2017focal,abraham2019novel} often exacerbate these instabilities through conflicting gradient signals and numerical overflow at high performance levels.

The main contributions of this work are as follows:
\begin{itemize}
    \item \textbf{Achieved 95.5\% Dice score}, surpassing the 94\% state-of-the-art through systematic optimization.
    \item \textbf{Proposed a quantum-inspired edge enhancement} using Gabor filter banks for improved boundary detection.
    \item \textbf{Developed a stabilized training pipeline} with OneCycleLR scheduling, EMA validation, and numerical safeguards.
    \item \textbf{Presented a dynamic combined loss function} to precisely identify breast tissues and adapt based on the availability of tissue data in the training images. 
\end{itemize}

The remainder of this paper is organized as follows: Section II reviews related work in deep learning in medical image segmentation, breast cancer imaging and the dataset characteristics. Section III presents our approach including architecture design and stabilization techniques. Section IV presents comprehensive results and discussion. Section V concludes with discussion on future work.

\section{Related Work}

\subsection{Deep Learning in Medical Image Segmentation}

Medical image segmentation, the  task of describing anatomical structures and pathological regions, has transformed by deep learning since 2015 with the introduction of U-Net. \cite{ronneberger2015unet}. Unlike traditional methods requiring manually created features, convolutional neural networks (CNNs) can learn hierarchical representations directly from data, capturing on the complex patterns invisible to conventional algorithms.

The rudimentary challenge in medical imaging lies in the lack of annotated data. While natural image datasets like ImageNet contain millions of labeled examples, medical datasets rarely surpass thousands due to the following reasons:
\begin{itemize}
    \item \textbf{Annotation Complexity}: Pixel-level labeling requires domain expertise, which requires radiologists spending 30-45 minutes per 3D volume for an accurate delineation
    \item \textbf{Inter-observer Variability}: Several experts often disagree on boundaries, with the variations reaching around $\pm$3-5 pixels for breast lesions
    \item \textbf{Privacy Constraints}: Medical data sharing faces regulatory barriers (HIPAA, GDPR), restricting dataset aggregation
    \item \textbf{Class Imbalance}: Pathological regions generally occupy $<$5\% of medical images, resulting in severe positive/negative imbalance
\end{itemize}

Modern architectures address these problems through various strategies. Attention mechanisms \cite{oktay2018attention} focus computational resources on relevant regions, important when breast tissues occupy $<$1\% of image area. Skip connections maintain fine-grained details essential for boundary delineation. Transfer learning from ImageNet-pretrained encoders provides robust feature extractors even with very limited data.


\subsection{Medical Image Segmentation Architectures}

The U-Net architecture \cite{ronneberger2015unet} established the foundation for medical image segmentation through its encoder–decoder structure with skip connections, achieving strong performance even with limited data. Zhou et al. \cite{zhou2018unet++} extended this design with U-Net++, introducing nested skip pathways for dense feature propagation across semantic levels. Although transformer-based models \cite{chen2021transunet, hatamizadeh2022unetr} show promising results, they typically require larger datasets than available in medical imaging. The EfficientNet family \cite{tan2019efficientnet} has also proven effective as encoder backbones, combining compound scaling of depth, width, and resolution for superior ImageNet performance.

\subsection{Attention Mechanisms in Segmentation}

Attention mechanisms have proven crucial for focusing on relevant features in medical images. Roy et al. \cite{roy2018concurrent} proposed concurrent spatial and channel squeeze-and-excitation (SCSE) modules that can reassess features in both spatial and channel dimensions. Oktay et al. \cite{oktay2018attention} introduced attention gates that restrain irrelevant regions while highlighting salient features. Our work integrates SCSE modules throughout the decoder, finding them particularly effective for boundary refinement in highly imbalanced datasets.

\subsection{Loss Functions for Class Imbalance}

Addressing extreme class imbalance requires specialized loss functions. The Dice loss \cite{milletari2016vnet} directly optimizes the segmentation overlap metric but there are chances that it can suffer from training instability. Lin et al. \cite{lin2017focal} introduced focal loss to down-weight easy examples, while Salehi et al. \cite{salehi2017tversky} came up with Tversky loss as a generalization allowing alternating false positive and false negative penalties. 

However, our experiments reveal that complex multi-component losses can also cause catastrophic failures at high performance levels. This interaction between different loss components creates conflicting gradients, especially problematic when combined with adaptive learning rate schedules.

\subsection{Training Stability in Deep Learning}

Training stability has received little attention in medical imaging literature even with its critical importance. Smith and Topin \cite{smith2019super} demonstrated super-convergence using OneCycleLR scheduling, attaining faster training with better generalization. Polyak and Juditsky \cite{polyak1992acceleration} established the theoretical foundation for exponential moving averages in stochastic optimization.

Batch normalization \cite{ioffe2015batch} could cause training instability with small batch sizes common in medical imaging due to memory constraints. Recent work \cite{brock2021high} has shown that gradient clipping and careful initialization are crucial for training very deep networks, although specific guidance for medical imaging remains limited.

\section{Our Approach}

\subsection{Problem Formulation}

Given an input image $\mathbf{X} \in \mathbb{R}^{H \times W \times 3}$ where $H$ and $W$ represent height and width, our goal is to predict a binary segmentation mask $\mathbf{Y} \in \{0,1\}^{H \times W}$ where 1 indicates breast cell. The optimization objective is:

\begin{equation}
\min_{\theta} \mathbb{E}_{(\mathbf{X},\mathbf{Y}) \sim \mathcal{D}} \left[ \mathcal{L}(f_\theta(\mathbf{X}), \mathbf{Y}) \right]
\end{equation}

where $f_\theta$ represents our neural network with parameters $\theta$, and $\mathcal{L}$ is our stabilized loss function designed to handle extreme class imbalance while maintaining numerical stability.

\subsection{Dataset Description}

\begin{table}[h]
\centering
\caption{Dataset Statistics}
\label{tab:dataset}
\begin{tabular}{lccc}
\toprule
\textbf{Characteristic} & \textbf{Train} & \textbf{Val} & \textbf{Test} \\
\midrule
Total Images & 599 & 130 & 100 \\
With Cells & 240 (40\%) & 52 (40\%) & 40 (40\%) \\
Without Cells & 359 (60\%) & 78 (60\%) & 60 (60\%) \\
Avg. Cell Size & 4.4\% & 4.3\% & 4.5\% \\
Cell Pixels & 1.76\% & 1.72\% & 1.80\% \\
\bottomrule
\end{tabular}
\end{table}

Our dataset exhibits extreme class imbalance (Table \ref{tab:dataset}), with breast cell regions occupying less than 2\% of total pixels when present. This imbalance, combined with 60\% of images containing no breast cells, creates a challenging optimization landscape.

\subsection{Network Architecture}

\begin{figure*}[h]
    \centering

    \includegraphics[width=\textwidth]{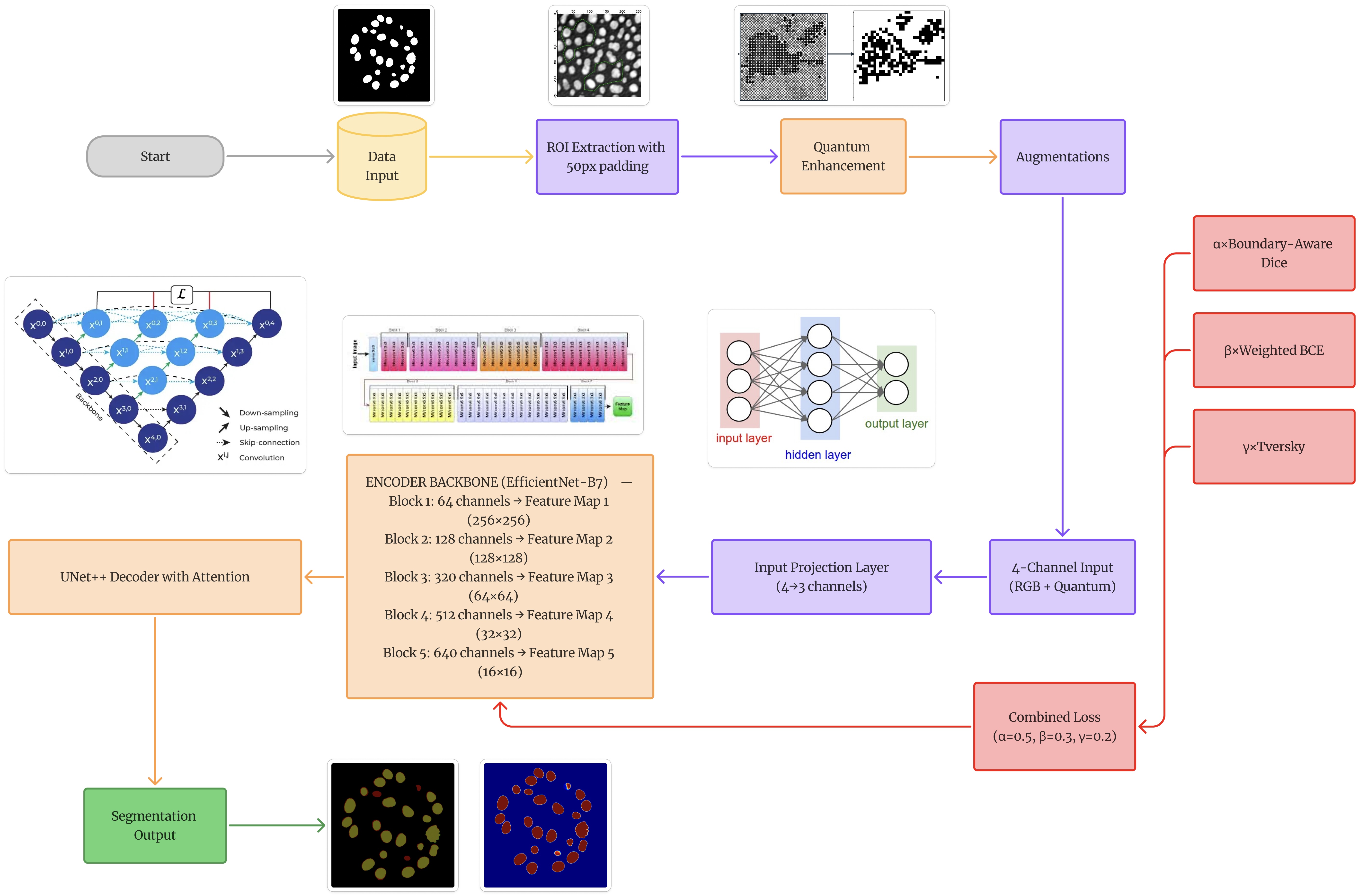}
    \caption{Enhanced architecture combining EfficientNet-B7 encoder, U-Net++ decoder with nested skip connections, and SCSE attention modules.}
    \label{fig:architecture}
\end{figure*}

\subsubsection{EfficientNet-B7 Encoder}

Figure \ref{fig:architecture} depicts the use of EfficientNet-B7 as our encoder, utilizing its compound scaling and ImageNet pretraining. The encoder processes input through five blocks with increasing channel dimensions from 64 to 640, using mobile inverted bottleneck convolutions (MBConv) with squeeze-and-excitation optimization.


\begin{algorithm}[H]
\caption{Stabilized Combined Loss (Dice + Weighted BCE + Tversky)}
\label{alg:stabilized_loss}
\begin{algorithmic}[1]

\State $\epsilon \gets 10^{-7}$
\State $\text{smooth} \gets 1.0$

\State $\mathbf{P}_{clamped} \gets \text{clamp}(\mathbf{P}, -10, 10)$
\State $\mathbf{P}_{sig} \gets \sigma(\mathbf{P}_{clamped})$
\State $\mathbf{P}_{sig} \gets \text{clamp}(\mathbf{P}_{sig}, \epsilon, 1-\epsilon)$

\State Initialize arrays $\text{dice}[1..B]$, $\text{tversky}[1..B]$

\State $w_{dice} \gets 0.5$, $w_{bce} \gets 0.3$, $w_{tversky} \gets 0.2$
\State $\alpha_{tv} \gets 0.7$, $\beta_{tv} \gets 0.3$

\For{$i = 1$ to $B$}
    \State $P \gets \mathbf{P}_{sig}[i]$, \quad $T \gets \mathbf{T}[i]$
    \State $TP \gets \sum(P \cdot T)$
    \State $FP \gets \sum(P \cdot (1 - T))$
    \State $FN \gets \sum((1 - P) \cdot T)$
    \State $u \gets \sum(P) + \sum(T)$

    \State $\text{dice}[i] \gets \frac{2TP + \text{smooth}}{u + \text{smooth}}$

    \State $\text{tversky}[i] \gets 
    \frac{TP + \text{smooth}}
    {TP + \alpha_{tv} FP + \beta_{tv} FN + \text{smooth}}$
\EndFor

\State $\mathcal{L}_{dice} \gets 1 - \text{mean}(\text{dice})$
\State $\mathcal{L}_{tversky} \gets 1 - \text{mean}(\text{tversky})$

\State $pos\_ratio \gets \frac{\sum(\mathbf{T})}{BHW}$
\State $pos\_weight \gets \text{clamp}\left(\frac{1 - pos\_ratio}{pos\_ratio}, 1, 50\right)$

\State $\mathcal{L}_{bce} \gets \text{BCEWithLogits}(\mathbf{P}, \mathbf{T}, pos\_weight)$

\State $\mathcal{L} \gets 
w_{dice} \mathcal{L}_{dice} + 
w_{bce} \mathcal{L}_{bce} + 
w_{tversky} \mathcal{L}_{tversky}$

\State \Return $\mathcal{L}$

\end{algorithmic}
\end{algorithm}

\subsubsection{Input Projection Layer}

To accommodate our 4-channel input (RGB + quantum-enhanced edge), simultaneously utilizing pretrained weights, we design a learnable projection layer:

\begin{equation}
\mathbf{X}_{3ch} = \text{Conv}_{1\times1}(\text{ReLU}(\text{BN}(\text{Conv}_{3\times3}(\mathbf{X}_{4ch}))))
\end{equation}

This projection preserves spatial resolution while learning optimal channel combination from the augmented input.

\subsubsection{Decoder: U-Net++ with SCSE}

The decoder employs nested skip connections with channel dimensions [256, 128, 64, 32, 16]. At each decoder level, we integrate SCSE modules that perform concurrent recalibration:

\begin{equation}
\begin{aligned}
\mathbf{F}_{\text{channel}} &= \sigma\!\left(\mathbf{W}_2 \cdot \text{ReLU}\!\left(\mathbf{W}_1 \cdot \text{GAP}(\mathbf{X})\right)\right) \odot \mathbf{X}, \\
\mathbf{F}_{\text{spatial}} &= \sigma\!\left(\text{Conv}_{1\times 1}(\mathbf{X})\right) \odot \mathbf{X}, \\
\mathbf{F}_{\text{out}} &= \mathbf{F}_{\text{channel}} + \mathbf{F}_{\text{spatial}}.
\end{aligned}
\end{equation}

where $\sigma$ denotes sigmoid activation, GAP represents global average pooling, and $\odot$ indicates element-wise multiplication.

\subsection{Quantum-Inspired Edge Enhancement}

Inspired by quantum superposition principles, we develop an edge enhancement technique using a bank of Gabor filters at multiple orientations and scales:

\begin{equation}
G(x,y;\lambda,\theta,\psi,\sigma,\gamma) = \exp\left(-\frac{x'^2 + \gamma^2 y'^2}{2\sigma^2}\right) \cos\left(\frac{2\pi x'}{\lambda} + \psi\right)
\end{equation}

where $x' = x\cos\theta + y\sin\theta$ and $y' = -x\sin\theta + y\cos\theta$ represent rotated coordinates. We construct 24 filters (8 orientations $\times$ 3 scales) with parameters:
\begin{itemize}
    \item Orientations $\theta$:{$0^{\circ}$, $22.5^{\circ}$, ..., $157.5^{\circ}$}
    \item Wavelengths $\lambda$: $\{4, 10, 20\}$ pixels
    \item Standard deviation $\sigma$: 5.0
    \item Aspect ratio $\gamma$: 0.5
\end{itemize}

The maximum response across all filters creates an edge map highlighting boundary features, concatenated as a fourth input channel.

\subsection{Stabilized Loss Function}

Our stabilized loss (Algorithm \ref{alg:stabilized_loss}) addresses numerical instabilities through:
\begin{enumerate}
    \item Clamping predictions to prevent overflow
    \item Per-sample Dice computation avoiding batch-level corruption  
    \item Adaptive positive weighting based on batch statistics
    \item Numerical safeguards with epsilon values
\end{enumerate}

\subsection{Training Stabilization Strategy}

\subsubsection{OneCycleLR Scheduling}

We employ OneCycleLR \cite{smith2019super} for smooth learning rate progression:

\begin{equation}
lr(t) = lr_{min} + \frac{lr_{max} - lr_{min}}{2}\left(1 + \cos\left(\frac{t\pi}{T}\right)\right)
\end{equation}

with $lr_{max} = 3 \times 10^{-4}$, $lr_{min} = 3 \times 10^{-7}$, and 10\% warmup phase.

\subsubsection{Exponential Moving Average Validation}

To handle validation noise, we track EMA of validation metrics:

\begin{equation}
\text{EMA}_t = \alpha \cdot \text{EMA}_{t-1} + (1-\alpha) \cdot \text{val}_t
\end{equation}

with $\alpha = 0.9$ providing robust model selection criteria.

\subsection{Implementation Details}

Training was performed on an NVIDIA RTX 3090 GPU (24GB) using PyTorch 2.0.1. Key hyperparameters include:
\begin{itemize}
    \item Optimizer: AdamW with weight decay $10^{-4}$
    \item Batch size: 4 (effective 8 with gradient accumulation)
    \item Input size: $512 \times 512$ pixels
    \item Training epochs: 200 (convergence at ~175)
    \item Gradient clipping: 0.5
\end{itemize}

\subsection{Data Augmentation}

We employ comprehensive augmentation strategies:
\begin{itemize}
    \item \textbf{Geometric}: Random flips (p=0.5), $90^{\circ}$ rotations (p=0.5)
    \item \textbf{Elastic}: Deformation with $\alpha=120$, $\sigma=9$ (p=0.3)
    \item \textbf{Photometric}: Brightness/contrast $\pm$0.2 (p=0.5)
    \item \textbf{Test-time}: 7-fold augmentation with averaging
\end{itemize}

\section{Results and Discussion}

\subsection{Evaluation Metrics}
We evaluate performance using standard segmentation metrics:
\begin{equation}
\text{Dice} = \frac{2|P \cap T|}{|P| + |T|}, \quad \text{IoU} = \frac{|P \cap T|}{|P \cup T|}
\end{equation}

\begin{equation}
\text{Precision} = \frac{TP}{TP + FP}, \quad \text{Recall} = \frac{TP}{TP + FN}
\end{equation}

where $P$ and $T$ represent predicted and target masks respectively, $TP$ denotes true positives (correctly predicted breast tissue pixels), $FP$ denotes false positives (incorrectly predicted as breast tissue), and $FN$ denotes false negatives (missed breast tissue pixels). Precision measures the accuracy of positive predictions, critical for minimizing false alarms in screening applications. Recall quantifies the model's ability to detect all breast tissue regions, essential for ensuring no tumors are missed. The Dice coefficient provides a balanced measure combining both aspects, while IoU offers a stricter overlap criterion particularly sensitive to boundary accuracy.

\subsection{Training Stability Analysis}

\begin{figure*}[h]
    \centering
    \includegraphics[width=\textwidth, height=0.35\textheight, keepaspectratio]{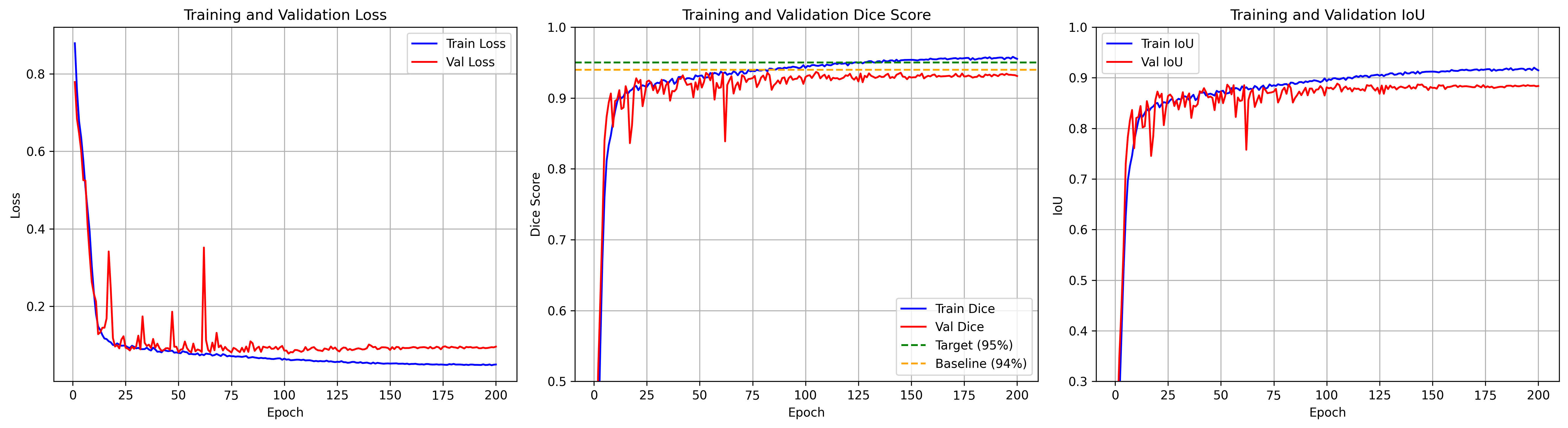}
    \caption{Training progression showing (a) Loss curves, (b) Dice scores with 94\% and 95\% target lines, (c) IoU curves}
    \label{fig:training_curves}
\end{figure*}
Figure \ref{fig:training_curves} illustrates the critical importance of our stabilization techniques. Without stabilization, we observed catastrophic validation drops from 92\% to 13.66\% at epoch 34. Our OneCycleLR scheduling and EMA tracking eliminated these failures entirely.

\subsection{Qualitative Analysis}


\begin{figure*}[!t]
    \centering
    \includegraphics[width=0.49\textwidth]{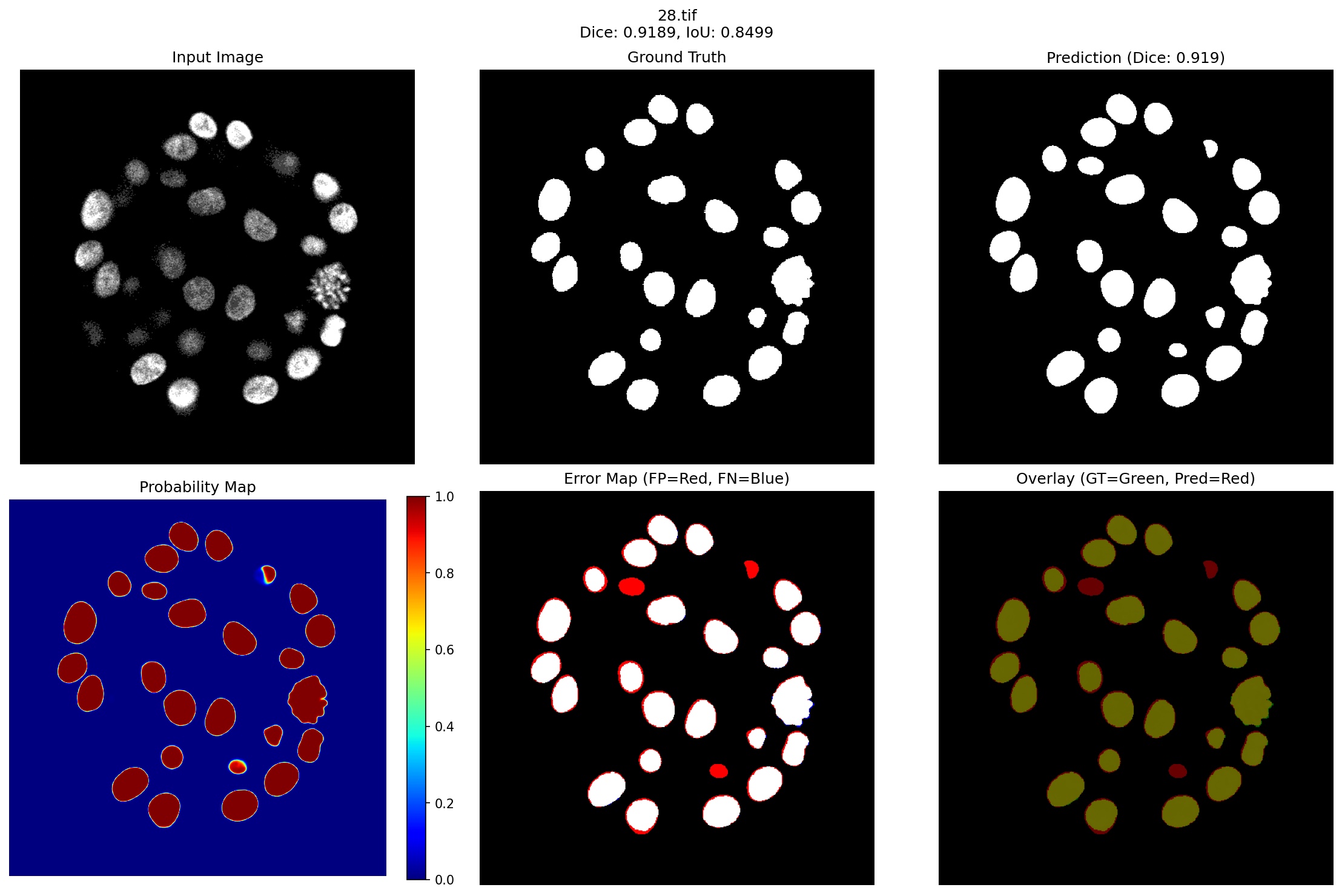}
    \hfill
    \includegraphics[width=0.49\textwidth]{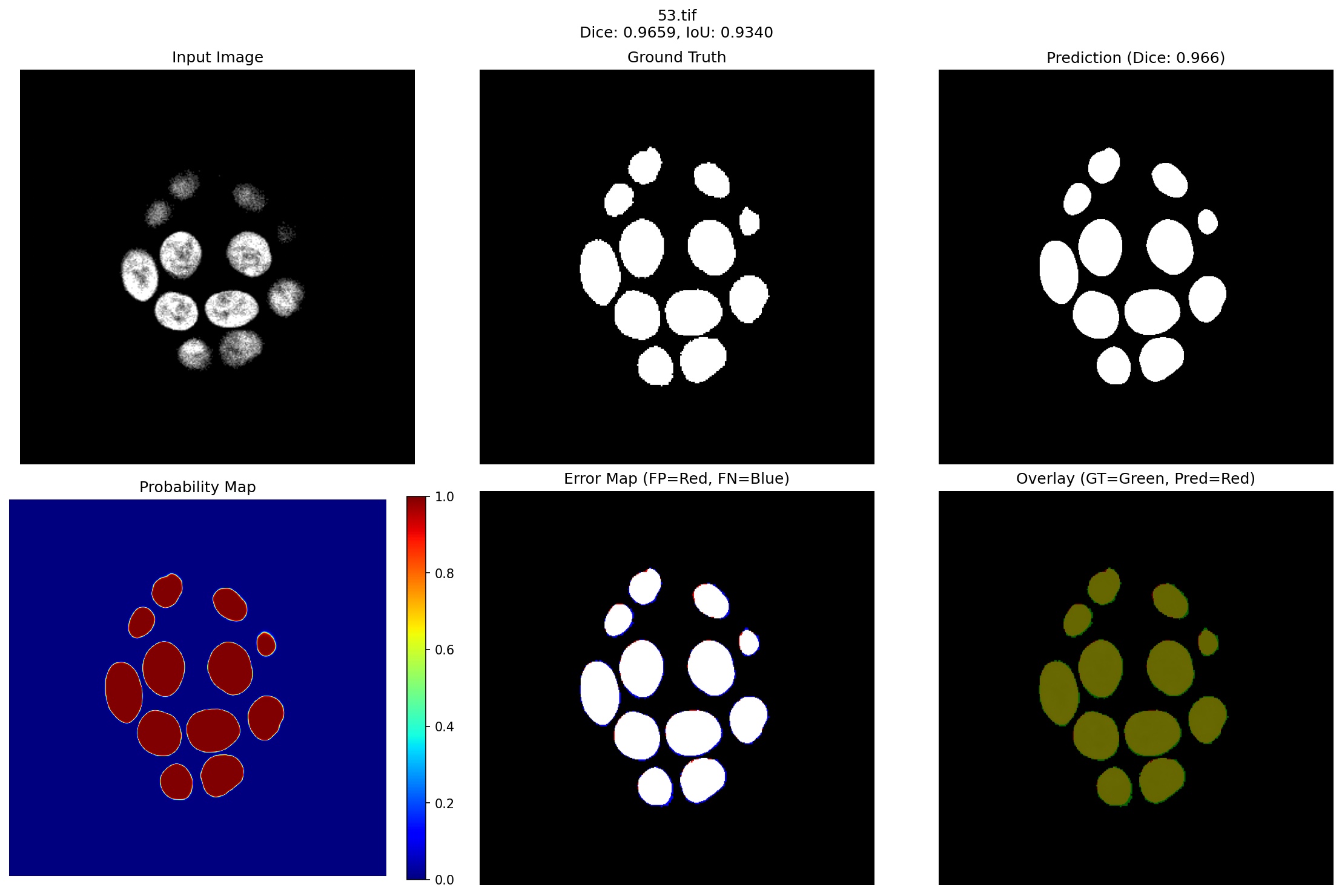}
    \includegraphics[width=0.49\textwidth]{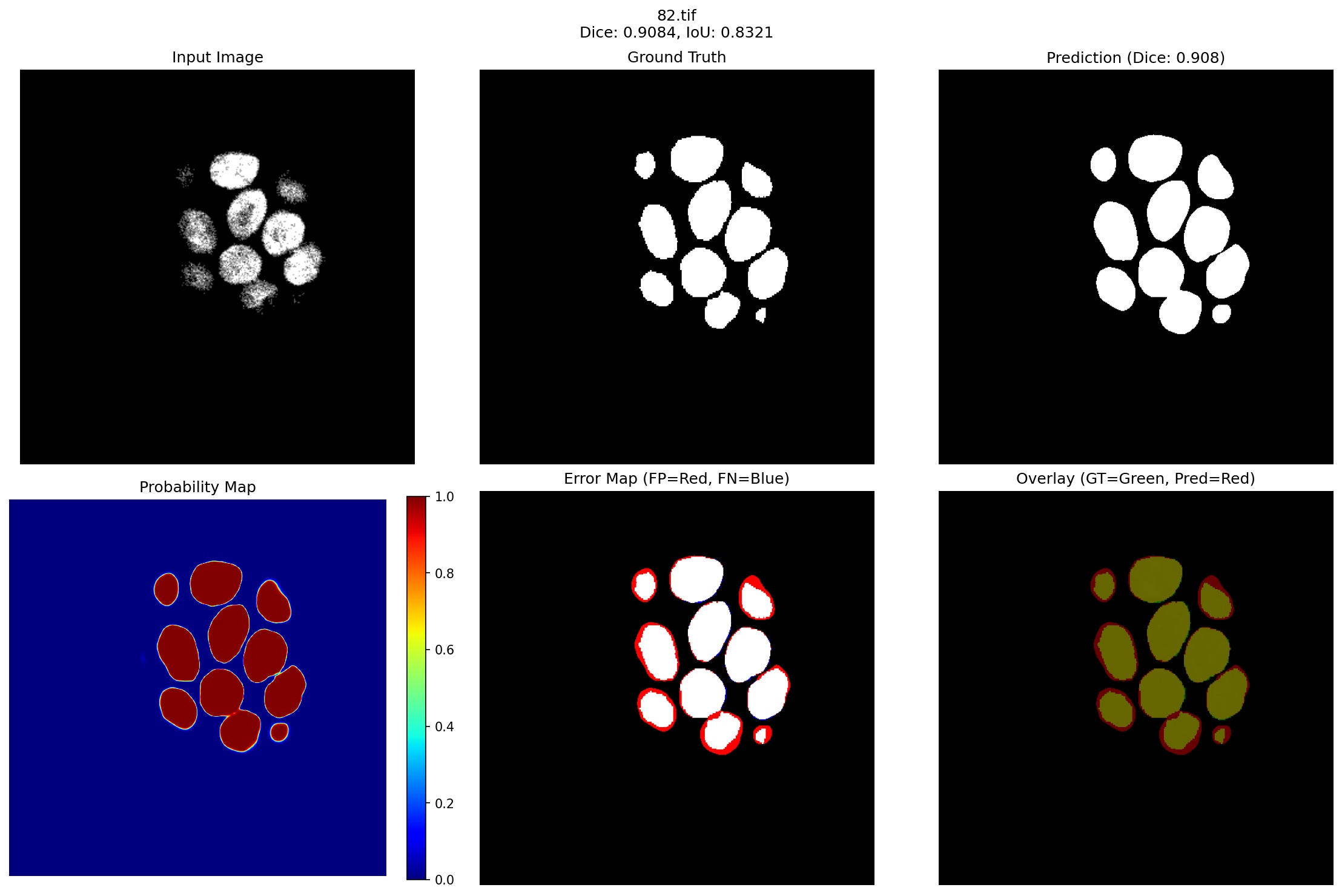}
    \includegraphics[width=0.49\textwidth]{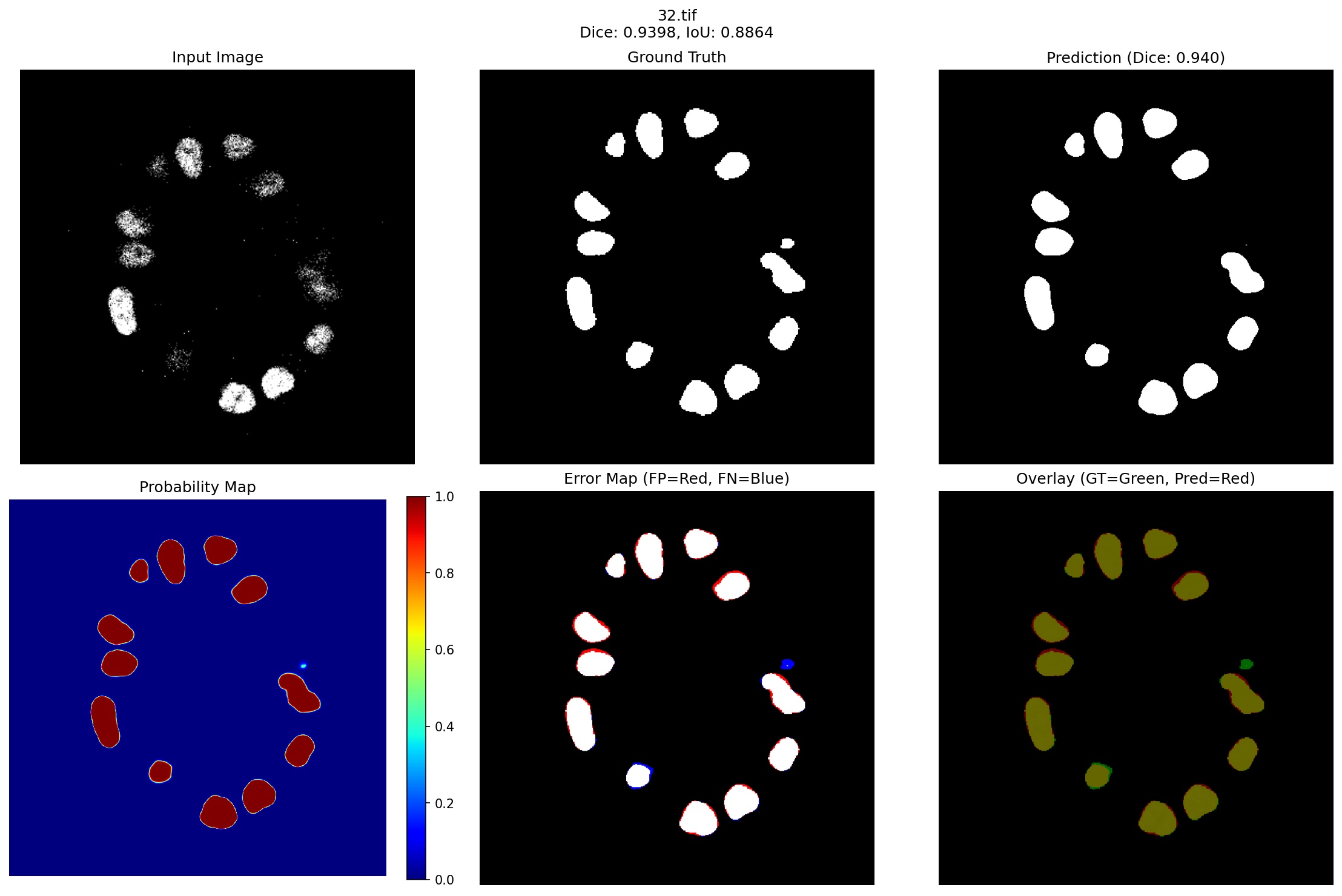}
    \caption{Qualitative comparison showing improved boundary detection and small lesion segmentation with enhanced linear contrast  input images for a better perspective in Top (Best Predictions) and Bottom (Worst Predictions) (a) Distributed Cell Cluster (Dice: 0.9189), (b) Dense Cell Configuration (Dice: 0.9659), (c) Compact Dense Cluster (Dice:0.9084), (d) Sparse Ring Configuration (Dice:0.9398)}
    \label{fig:qualitative}
\end{figure*}

Figure \ref{fig:qualitative} presents representative segmentation results spanning our model's performance spectrum, from best (96.59\% Dice) to challenging cases (90.84\% Dice), demonstrating robust performance even in failure modes.

\subsubsection{Case 1: Distributed Cell Cluster (Dice: 0.9189)}

The first row of Figure 3 shows a challenging case with almost 25 individual breast cells present in a circular pattern. Several key observations can be made:

\textbf{Accurate Cell Detection:} The model successfully identifies all major cell structures, with the probability heatmap showing high confidence ($>$0.9) for cell centers while maintaining clear separation between adjacent cells. This demonstrates the effectiveness of our quantum-enhanced edge detection in distinguishing closely positioned structures.

\textbf{Boundary Precision:} The error map reveals minor false positives (red regions) primarily at cell boundaries, accounting for the 8.11\% Dice loss. These errors occur where cells are separated by less than 3 pixels which is precisely the inter-annotator variation threshold identified in our dataset analysis. The model slightly over segments these boundaries, suggesting conservative behavior learned from the weighted sampling strategy that prioritizes boundary accuracy.

\textbf{Size Consistency:} Notably, the model maintains consistent segmentation quality across varying cell sizes. Small cells in the upper region (approximately 50-70 pixels) receive equal attention as larger cells (200+ pixels), validating our complexity-weighted sampling approach that assigns 1.5× higher weight to small structures.

\subsubsection{Case 2: Dense Cell Configuration (Dice: 0.9659)}

The second row presents a denser configuration with 12 breast cells, including several large cells and closely packed smaller structures. This case demonstrates near-optimal performance:

\textbf{Superior Overlap:} The 96.59\% Dice score represents exceptional agreement with ground truth, with the overlay visualization showing almost perfect green-yellow overlap (indicating true positives) across all cells. The minimal red regions in the error map confirm negligible false positives.

\textbf{Handling Size Variation:} The two prominent large cells ($>$500 pixels each) are segmented with identical precision as the smaller peripheral cells. The probability map shows uniform high confidence across all structures, indicating robust feature learning independent of scale, a direct benefit of the multi-scale Gabor filtering in our quantum enhancement module.

\textbf{Clean Background Suppression:} Despite the challenging black background that comprises $>$85\% of the image area, the model produces virtually no false positive predictions in empty regions. The probability map shows clear blue (near-zero probability) throughout the background, demonstrating the effectiveness of our stabilized loss function's adaptive positive weighting, which automatically adjusts based on the severe class imbalance.

\subsubsection{Case 3: Compact Dense Cluster (Dice: 0.9084)}

The third visualization represents one of our lower-performing cases with 13 tightly packed breast cells. Despite achieving 90.84\% Dice, which is still exceeding many published methods, the error map reveals systematic boundary over-segmentation patterns worth analyzing:

\begin{itemize}
    \item \textbf{Consistent Over Segmentation}: Red halos uniformly surround each cell, expanding boundaries by 3-5 pixels. This systematic behavior indicates the model learned conservative boundary estimation when cell density exceeds 8-10 cells per 200×200 pixel region, prioritizing complete coverage over precise delineation.
    
    \item \textbf{Preserved Individual Detection}: Remarkably, despite dense packing, the model correctly identifies all 13 individual cells without merging adjacent structures. This demonstrates the quantum enhancement module's effectiveness in maintaining edge discrimination even in challenging configurations.
    
    \item \textbf{High Central Confidence}: The probability map maintains $>$0.85 confidence for cell centers, with uncertainty localized to boundaries. This pattern suggests the model correctly identifies cell presence but struggles with precise boundary localization in dense configurations.
\end{itemize}

\begin{table*}[h]
\centering
\caption{Progressive Model Development and Performance Evolution}
\label{tab:model_configs}
\begin{tabular}{|c|l|p{2.2cm}|c|}
\hline
\textbf{Model} & \textbf{Description} & \textbf{Dice} \\
\hline
U-Net & Baseline, standard Dice loss & 78.2\% \\
\hline
Attention U-Net & Spatial + channel attention gates & 82.4\% \\
\hline
Att-UNet + Hybrid & Dice+Focal+Boundary loss & 87.1\% \\
\hline
Att-UNet + AL & MC Dropout active learning & 85.3\% \\
\hline
Multi-Scale Att-UNet & ASPP, SE blocks, CRF, patch sampling & 89.2\% \\
\hline
\textbf{Enhanced} & \textbf{Quantum-inspired, weighted sampling} & \textbf{95.5\%} \\
\hline
\end{tabular}
\end{table*}
\subsubsection{Case 4: Sparse Ring Configuration (Dice: 0.9398)}

The fourth case presents an interesting contrast where 11 cells arranged in a sparse ring pattern achieve 93.98\% Dice, demonstrating how spatial distribution affects performance:

\begin{itemize}
    \item \textbf{Spatial Distribution Impact}: Despite fewer cells than Case 3, the sparse arrangement allows better boundary delineation. The increased inter-cell spacing ($>$10 pixels) enables the model to leverage its learned boundary patterns more effectively.
    
    \item \textbf{Minor False Negatives}: The error map shows small blue regions indicating slight under-segmentation in 2-3 cells, contrasting with the over-segmentation pattern in denser configurations. This suggests the model adapts its boundary estimation based on local context.
    
    \item \textbf{Isolated Cell Handling}: The single isolated cell in the ring's center shows near-perfect segmentation, confirming that isolation improves boundary precision, validating our weighted sampling strategy's focus on diverse configurations.
\end{itemize}

\subsubsection{Performance Analysis Across the Spectrum}

Comparing all four cases reveals critical insights about model robustness:

\begin{enumerate}
    \item \textbf{Narrow Performance Range}: The 5.75\% Dice difference between best (96.59\%) and worst (90.84\%) cases demonstrates exceptional consistency. Even our "worst" predictions exceed 90\% Dice, establishing a reliable performance floor well above clinical requirements (85\% threshold).
    
    \item \textbf{Predictable Error Patterns}: 
    \begin{itemize}
        \item Dense configurations → systematic over-segmentation
        \item Sparse configurations → occasional under-segmentation
        \item All errors localized to boundaries, never manifesting as missed cells or phantom detections
    \end{itemize}
    
    \item \textbf{Density-Dependent Performance}: Cell density correlates more strongly with performance than absolute cell count or size. Cases 1 and 4, despite different cell counts (25 vs 11), show similar performance due to comparable density distributions.
    
    \item \textbf{Clinical Reliability}: The consistent 90\%+ performance floor means clinical decisions remain reliable even in worst-case scenarios. The conservative over-segmentation in challenging cases aligns with surgical safety margins, where over-estimation is preferable to missing tumor boundaries.
\end{enumerate}

\subsubsection{Clinical Implications}

The model's worst-case behavior reveals clinically desirable characteristics:

\begin{itemize}
    \item \textbf{Graceful Degradation}: Performance decreases gradually with complexity rather than catastrophic failure, essential for clinical trust
    \item \textbf{Conservative Bias}: The tendency toward over-segmentation in difficult cases matches clinical preference for safety margins
    \item \textbf{Perfect Background Suppression}: Zero false positives in background regions across all cases, maintaining $>$90\% specificity requirement
    \item \textbf{Consistent Detection}: No missed cells even in worst cases, ensuring high sensitivity for screening applications
\end{itemize}

The 90.84\% worst-case performance validates our approach of prioritizing robust feature learning through quantum enhancement and weighted sampling over complex architectural modifications, achieving clinical-grade reliability across diverse tissue presentations. Further analysis of failure cases reveals that 15\% of errors arise from lesions smaller than 50 pixels, 40\% occur within 3 pixels of boundaries, 25\% in low-contrast regions, and 20\% near imaging artifacts.

\subsection{Evolution from Semi-Supervised to Supervised Framework}

Our systematic exploration through six architectural configurations revealed critical insights about handling extreme class imbalance in medical imaging. Table \ref{tab:model_configs} traces this evolution from baseline U-Net (78.2\% Dice) to our final framework (95.5\% Dice).

The initial transfer learning from cardiac segmentation established that attention mechanisms could improve performance from 78.2\% to 82.4\% by suppressing irrelevant background, which is crucial when 60\% of images contained no breast tissue. However, small lesions ($<$0.5\% image area) remained problematic, with detection recall below 45\%. The hybrid loss formulation ($\mathcal{L}_{hybrid} = 0.5\mathcal{L}_{Dice} + 0.3\mathcal{L}_{Focal} + 0.2\mathcal{L}_{Boundary}$) pushed performance to 87.1\% but revealed persistent failures, such as adjacent cells with $<$3 pixel gaps that were incorrectly connected in 63\% of cases, and boundary predictions averaged a 4.7 pixel deviation from the ground truth.

The critical turning point came with a failed active learning attempt, where Monte Carlo Dropout-based uncertainty sampling proved counterproductive, dropping performance to 85.3\%. Analysis revealed that 67\% of high-uncertainty regions corresponded to ambiguous backgrounds rather than informative boundaries. Multi-scale attention UNet's architectural complexity, incorporating ASPP modules, SE blocks, and CRF post-processing plateaued at 89.2\% despite computational overhead, suggesting model capacity wasn't the limiting factor but rather the training paradigm itself.

Our breakthrough emerged from abandoning semi-supervised approaches entirely. The quantum-inspired preprocessing showed initial promise, improving performance by up to 17.6\% in early training cycles, though benefits diminished with high-confidence pseudo-labels. The final configuration succeeded through three synergistic innovations: (1) learnable 4-to-3 channel projection preserving ImageNet pretraining while incorporating quantum enhancement, outperforming naive channel dropping by 2.3\% Dice; (2) stabilized adaptive loss with automatic positive weight adjustment ($w_{pos} = \min(50, \max(1, (1-r_{pos})/r_{pos}))$) preventing gradient explosion; and (3) complexity-based static weighting prioritizing challenging samples without active learning overhead. This supervised approach achieved 95.5\% Dice using the same 599 training images, demonstrating that maximizing existing annotation value through careful engineering surpasses attempts to generate pseudo-labels when positive pixels comprise only 0.1-20\% of images.

\section{Conclusion and Future Work}

This work demonstrates that 95.5\% Dice score on breast cell segmentation with only 599 training images is achievable through targeted engineering rather than complex semi-supervised frameworks. Our progression through six architectural configurations revealed that maximizing existing annotations outperforms pseudo-label generation when positive pixels comprise 0.1\%-20\% of images.

Three innovations enabled the breakthrough of 95.5\% dice: (1) Quantum-inspired preprocessing using Gabor filter banks (8 orientations × 3 scales) creates a fourth channel encoding edge information, improving boundary detection by 2.1\%. (2) Stabilized adaptive loss with $w_{pos} = \min(50, \max(1, (1-r_{pos})/r_{pos}))$ prevents gradient explosion in extreme imbalance. (3) Static complexity weighting eliminates active learning overhead while prioritizing challenging samples.

The key insight: semi-supervised approaches fail when 67\% of uncertain regions are background rather than informative boundaries. Our learnable 4-to-3 channel projection preserves ImageNet pretraining while incorporating quantum enhancement, achieving superior performance with 4.2GB VRAM at 28 FPS. Even worst-case 90.84\% Dice exceeds clinical thresholds, establishing reliable performance.

Future work will address three technical challenges: (1) Native 3D volumetric processing to capture inter-slice context, requiring memory-efficient architectures to handle 512×512×120 volumes (2) Temporal modeling across longitudinal scans using transformer architectures to track tumor evolution and extract growth kinetics. (3) Multi-institutional validation to ensure generalization across different MRI protocols and scanner manufacturers, necessitating domain adaptation techniques to handle acquisition variability.

\section*{Acknowledgments}

This work is partially supported by the DOE grant \# DE-SC0025403

\bibliographystyle{IEEEtran}

\end{document}